\documentclass[runningheads]{llncs}
\usepackage{graphicx}

\usepackage{tikz}
\usepackage{comment}
\usepackage{amsmath,amssymb} 
\usepackage{color}
\usepackage{booktabs}
\usepackage{multirow}
\usepackage{multicol}
\usepackage{amsmath,bm}

\usepackage{soul}
\usepackage[accsupp]{axessibility}  

\usepackage[capitalize]{cleveref}
\crefname{section}{Sec.}{Secs.}
\Crefname{section}{Section}{Sections}
\Crefname{table}{Table}{Tables}
\crefname{table}{Tab.}{Tabs.}

\begin{document}
\pagestyle{headings}
\mainmatter
\def\ECCVSubNumber{4879}  

\title{Open-world Semantic Segmentation via Contrasting and Clustering Vision-Language Embedding} 


\titlerunning{ViL-Seg}
%
\author{Quande Liu\inst{1}
\and
Youpeng Wen\inst{2} \and
Jianhua Han\inst{3} \and
Chunjing Xu\inst{3} \and
Hang Xu\inst{3}$^\dagger$ \and
Xiaodan Liang\inst{2}$^\dagger$
}

\authorrunning{Q. Liu et al.}
%
\institute{The Chinese University of Hong Kong\\ \email{qdliu@cse.cuhk.edu.hk}\and
Shenzhen Campus of Sun Yat-sen University \\
\email{wenyoupeng0@outlook.com, xdliang328@gmail.com}\and
Huawei Noah's Ark Lab\\
\email{\{hanjianhua4,xuchunjing,xu.hang\}@huawei.com}
}
\maketitle

\begin{abstract}
To bridge the gap between supervised semantic segmentation and real-world applications that acquires one model to recognize arbitrary new concepts, recent zero-shot segmentation attracts a lot of attention by exploring the relationships between unseen and seen object categories, yet requiring large amounts of densely-annotated data with diverse base classes. In this paper, we propose a new open-world semantic segmentation pipeline that makes the first attempt to learn to segment semantic objects of various open-world categories without any efforts on dense annotations, by purely exploiting the 
image-caption data that naturally exist on the 
Internet.  Our method,  \textbf{Vi}sion-\textbf{l}anguage-driven Semantic \textbf{Seg}mentation (ViL-Seg), employs an image and a text encoder to generate visual and text embeddings for the image-caption data, with two core components that endow its segmentation ability: First, the image encoder is jointly trained with a vision-based contrasting and a cross-modal contrasting, which encourage the visual embeddings to preserve both fine-grained semantics and high-level category information that are crucial for the segmentation task. Furthermore, an online clustering head is devised over the image encoder, which allows to dynamically segment the visual embeddings into distinct semantic groups such that they can be classified by comparing with various text embeddings to complete our segmentation pipeline. Experiments show that without using any data with dense annotations, our method can directly segment objects of arbitrary categories, outperforming zero-shot segmentation methods that require data labeling on three benchmark datasets.
\let\thefootnote\relax\footnotetext{$^\dagger$ Corresponding authors.}
\end{abstract}

\section{Introduction}





\begin{figure}[tb]
		\begin{center}

\includegraphics[ width=0.9\linewidth]{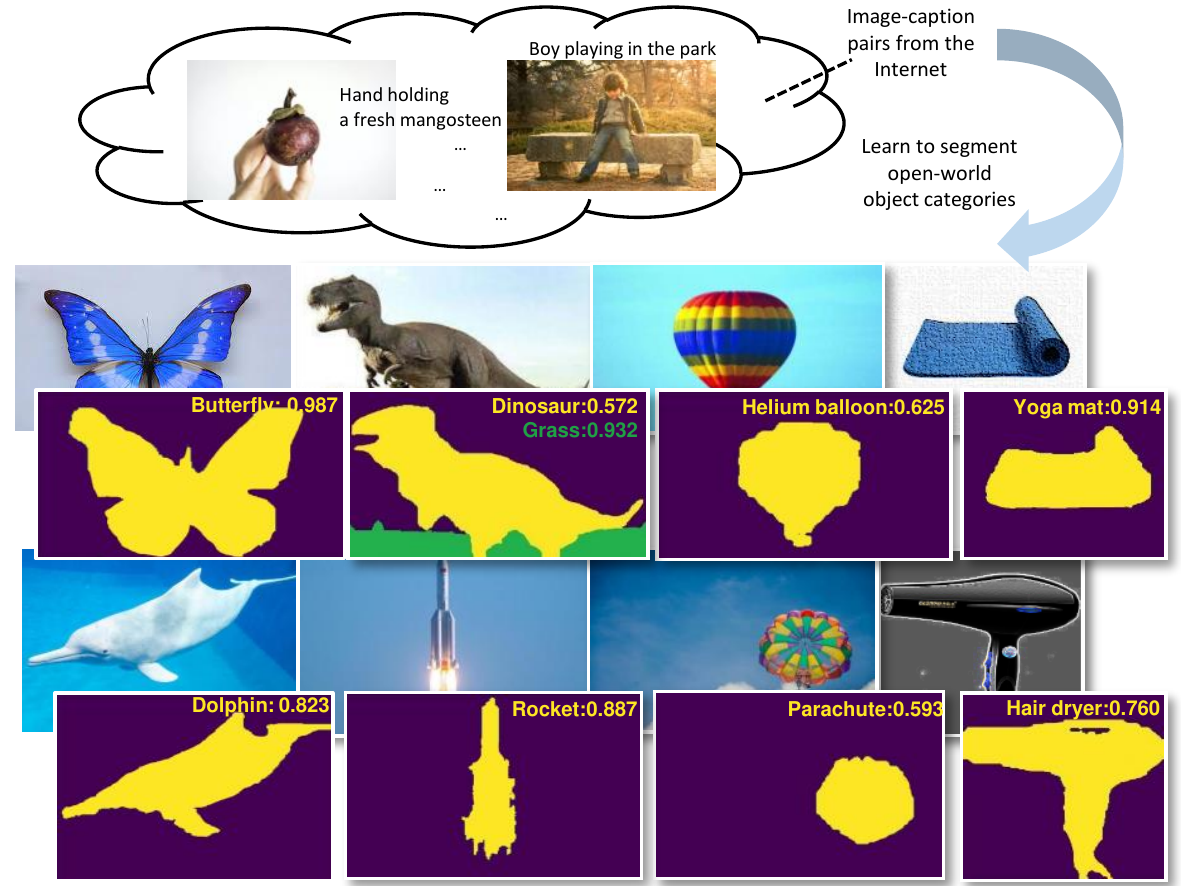}

		\end{center}
		\vspace{-4mm}
		\caption{By purely utilizing the image-caption pairs from the Internet (without using any data with dense annotations), ViL-Seg is able to segment various object categories in the open world even though they are never labeled in existing segmentation datasets.}
		\vspace{-5mm}
		\label{fig:intro_figure}
\end{figure}

As a crucial problem in computer vision, semantic segmentation~\cite{minaee2021image} aims to assign a class label to each pixel in the image. Most existing semantic segmentation methods~\cite{ronneberger2015u,chen2017deeplab,lin2017refinenet,long2015fully,zhao2017pyramid} are only capable of segmenting base categories appearing in the  training dataset.
However, the number of object classes in existing semantic segmentation datasets~\cite{everingham2010pascal,mottaghi2014role,caesar2018coco} is limited due to the costly pixel-wise annotations, e.g., PASCAL VOC~\cite{everingham2010pascal} with 20 categories and COCO Stuff~\cite{caesar2018coco} with 183 categories, which is far away from the number of object categories that exist in reality.
The usual way to increase the category number is by  annotating more images of the novel categories, which, however, not only requires tremendous human labeling efforts but also faces difficulty to collect enough samples given the extremely large class number in open-world~\cite{gupta2019lvis}.


Recently, zero-shot segmentation methods~\cite{xian2019semantic,bucher2019zero,gu2020context,cheng2021sign} have been proposed to generalize the semantic segmentation model to unseen classes by leveraging the word embeddings to discover the implicit relationship between base and novel classes.
However, since all these methods rely on the \textit{training on a specific dataset containing some base classes}, the developed segmentation model would be biased towards either seen classes or the training scenes~\cite{gu2020context}, which will hurt the segmentation performance on novel classes and the transfer ability to other datasets in real-world applications.

%

Inspired by the recent advance of vision-language pre-training methods~\cite{radford2021learning,lin2021m6}, we aim to learn a model that can segment various object categories in open-world by purely leveraging  the  vision-language  data  that exists naturally on  the  Internet (cf. Fig.~\ref{fig:intro_figure}).
Compared with traditional manually-annotated datasets, image-caption data from the Internet~\cite{thomee2016yfcc100m,changpinyo2021conceptual} is much easier to collect and needs no more costly human labeling process.
Besides, given the tremendous data resources on the Internet, these data can easily scale up to tens or hundreds of millions level and greatly increase the diversity of object categories~\cite{neilchen2013}, which paves the way for the model to handle object classes that are never labeled in existing datasets but exist in reality. 
%
%
Recently, there have been some studies~\cite{zareian2021open,gu2021openvocabulary} to exploit the large-scale vision-language data to solve some downstream tasks, such as image classification~\cite{radford2021learning} or captioning~\cite{su2019vl}. Zareian et al.~\cite{gu2021openvocabulary} also proposed to leverage the cross-modal data to address unseen-class object detection problem by distilling the knowledge from a pre-trained zero-shot classification model into an object detector.
However, how to leverage these web-based image-caption data to address the semantic segmentation problem for open-world object categories  remains unsolved, which is also highly challenging given that the caption only contains a global semantic description for the image which is insufficient for the segmentation task that requires dense semantic understanding.

In this paper, we present Vision-language-driven Semantic Segmentation (ViL-Seg), a new open-world annotation-free semantic segmentation pipeline that makes the first attempt  to  learn  to  segment  semantic objects  of  various open-world categories by purely exploiting the vision-language data from the Internet. 
In detail, ViL-Seg utilizes an image encoder and a text encoder to generate visual and text embeddings for two different modalities (i.e., image and caption).
To preserve the fine-grained semantics and high-level category information which are two key properties for the visual embeddings in segmentation task, the image encoder has been trained under the supervision of two complementary objectives, i.e., a) a vision-based contrasting by comparing global and local image patches to learn local to global correspondence; b) a cross-modal contrasting to exploit the category information from natural language supervision.
Furthermore, an online clustering head is further designed over the image encoder, which segments the fine-grained visual embeddings into distinct semantic groups such that they can be classified by comparing the alignment with text embeddings of various open-world object categories. This online clustering design also makes the training and inference of ViL-Seg end-to-end.

Our main contributions are summarized as follows:
\begin{itemize}
    \item We present Vision-language-driven Semantic Segmentation (ViL-Seg), which to our knowledge is the first attempt to use the image-caption pairs from the Internet to learn to segment objects of various open-world categories without using any densely-annotated data.
    
    
    \item To explore the segmentation-related knowledge from image-caption data, ViL-Seg employs two complementary contrastive objectives to promote the quality of visual embeddings, with an online clustering head to dynamically divide the visual embeddings into different semantic regions.  Both the training and inference of ViL-Seg are performed end-to-end.

   
    \item Experiments show that without using any data with dense annotations, our ViL-Seg can segment various open-world object  categories, and outperform state-of-the-art zero-shot segmentation methods that require data labeling on three benchmark datasets, e.g., 5.56\% mIoU increase on PASCAL VOC.
\end{itemize}

\section{Related Work}
\subsection{Zero-shot Semantic Segmentation.}

Zero-shot semantic segmentation~\cite{bucher2019zero} denotes segmenting unseen categories without training with any instances of them. 
For the past few years, some methods~\cite{kato2019zero,li2020consistent} have been proposed via learning word embeddings between seen and unseen categories.
For instance, SPNet~\cite{xian2019semantic} utilizes a generator to generate synthetic features from word embedding to match the corresponding vision features, while ZS3Net~\cite{bucher2019zero} projects visual semantic embedding into class probability via a fixed word embedding matrix of different classes.
To mitigate the seen categories' bias in SPNet and the model collapse problem in ZS3Net, CaGNet~\cite{gu2020context} proposes a contextual module to generate more diverse and context-aware word embeddings.
Based on these methods, SIGN~\cite{cheng2021sign} further adopts and improves standard positional encoding to integrate spatial information of feature level and proposes annealed self-training  to assign different importance to pseudo labels according to their confidence.

There are also several works~\cite{geng2020recent,perera2020generative,oza2019c2ae} concentrating on the open-set recognition problem~\cite{scheirer2012toward}, which aim to distinguish whether the sample is from novel classes without providing a specific unseen category name. 
A variety of works on unsupervised semantic segmentation~\cite{van2021unsupervised,hwang2019segsort,ye2019unsupervised} also tend to learn dense semantic representations without using segmentation labels.
However, these methods can only provide semantic groups by using clustering methods like K-Means~\cite{kodinariya2013review} as post-processing on the network features, yet cannot provide the category name for each semantic group.
Different from these methods, by exploiting the vision-language data from the internet~\cite{changpinyo2021conceptual}, our method is capable of predicting the class name for each image pixel without using any data with dense annotations. 

\subsection{Vision-language Pre-training.}

Vision-language pre-training~\cite{li2020oscar,jia2021scaling,li2021unimo,huang2021seeing,wang2021simvlm} with massive image-text pairs from the Internet has attracted more and more attention in recent years. By using contrastive pre-training to predict the correct pairs of image and text samples, CLIP~\cite{radford2021learning} achieves competitive results compared with the fully-supervised baseline on several downstream classification tasks. 
Some works \cite{lin2021m6,chen2020uniter} also introduce language-modeling-like objectives, including masked language/region modeling, image captioning and text-denoising to further improve the performance of vision-language models.
Moreover, several methods~\cite{huo2021wenlan,su2019vl} adopt a pre-trained object detector  to obtain a sequence of object embeddings as the visual features.

Very recently, some studies~\cite{zareian2021open,gu2021openvocabulary,xie2021zsdyolo} have proposed to leverage the pre-trained vision-language model to address the open-vocabulary object detection task, which aims at training a model to detect any object from a given vocabulary of classes.
Zareian et al.~\cite{zareian2021open} propose to learn a vision to language (V2L) layer during pre-training, and utilize it to initialize a Faster-RCNN model.
ViLD\cite{gu2021openvocabulary} distills the knowledge from a pre-trained zero-shot classifier into a two-stage detector.
Based on ViLD, ZSD-YOLO\cite{xie2021zsdyolo} further expands the thought of distillation into YOLOv5~\cite{glenn_jocher_2020_4154370}. 
There are also several studies~\cite{pakhomov2021segmentation,xu2021simple} that tend to leverage the vision-language models, e.g., CLIP, to reduce the annotation cost in semantic segmentation task. However, these studies either rely on the annotated data on seen classes for training~\cite{xu2021simple}, or can only support unsupervised segmentation that simply separates the image pixels into variant semantic clusters without providing the corresponding class labels~\cite{pakhomov2021segmentation}. In contrast, we aim to develop an complete semantic segmentation  pipeline that can segment various open-world objects, by purely utilizing the image-caption data from the internet without using any densely-annotated  data.

\section{Method}
\begin{figure*}[t!]
		\begin{center}
            \includegraphics[ width=0.95\linewidth]{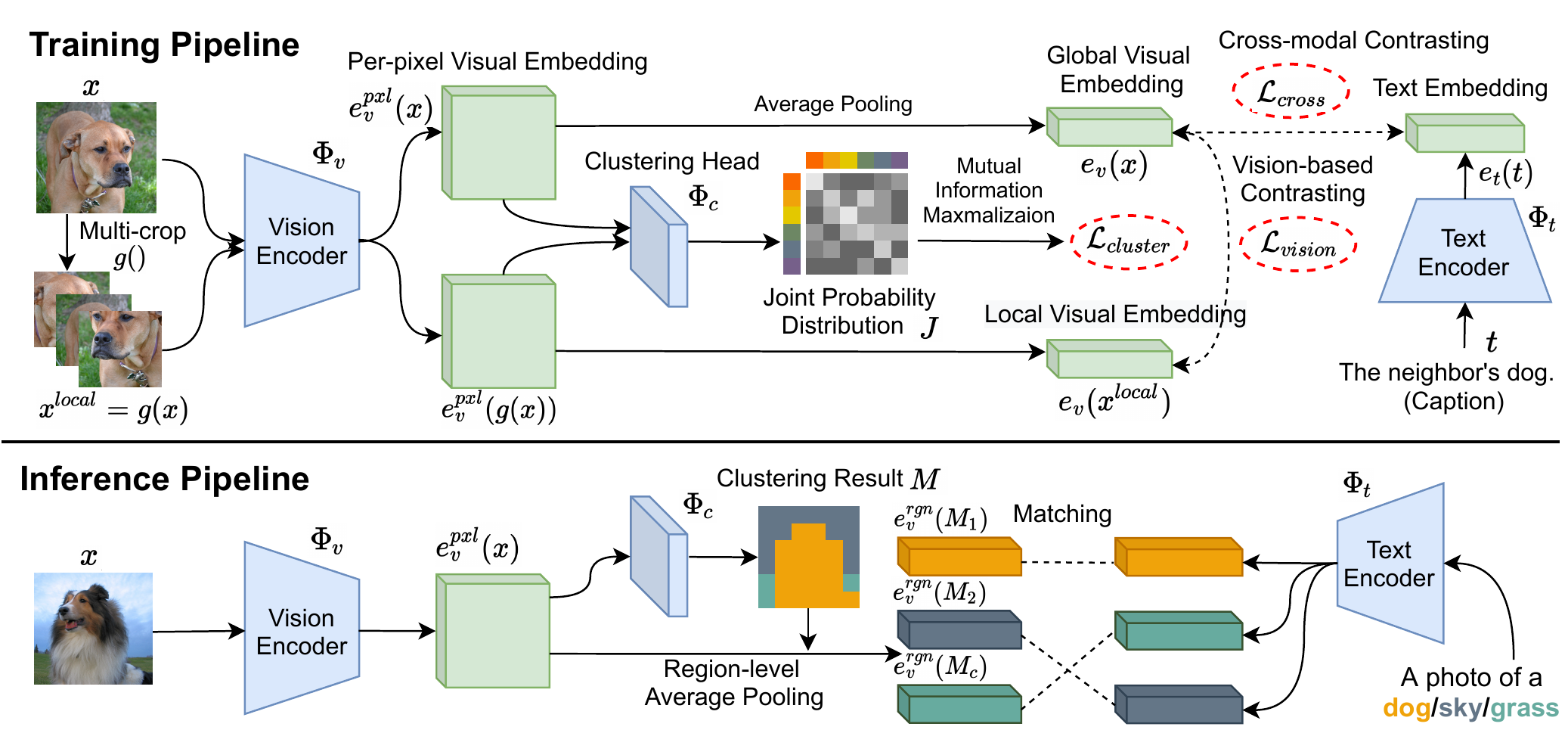}
		\end{center}
		\vspace{-4mm}
		\caption{Overall architecture of ViL-Seg. The image encoder is trained with two complementary objectives, i.e., the vision-based and cross-modal contrastive losses, aiming to promote the fine-grained semantics and high-level category information in the visual embeddings. Besides, an online clustering head is built over the image encoder to segment the pixel-wise visual embeddings into distinct semantic groups, which are trained with mutual information maximization. During inference, the segmentation is performed by comparing the feature pooled from each clustered region with different word embeddings. Both the training and inference are performed end-to-end.}
        \vspace{-4mm}
		\label{fig:framework}
\end{figure*}

Fig.~\ref{fig:framework} overviews our proposed Vision-language-driven Semantic Segmentation (ViL-Seg) method. In this section, we first briefly introduce its framework and training objective in Sec.~\ref{sec:framework}. Then, we describe the two complementary contrastive learning strategies which are used to enhance the visual embeddings in Sec.~\ref{sec:vision-cross-contrasting}, and present how to segment per-pixel visual embeddings into different semantic groups with the online clustering head in Sec.~\ref{sec:clustering}.

\subsection{ViL-Seg Framwork} 
\label{sec:framework}
The base of ViL-Seg is a vision encoder $\Phi_v$ and a text encoder  $\Phi_t$ to embed the image and its caption from the paired web data. We denote $e_v \in \mathbb{R}^{D}$ as the extracted global visual feature, $e_v^{pxl} \in \mathbb{R}^{HW \cdot D}$ as the per-pixel visual embeddings, e.g., embeddings before last pooling layer; and denote $e_t\in \mathbb{R}^{D}$ as the encoded text feature. To perform image segmentation 
over this framework, we also construct an online clustering head $\Phi_c$ over the image encoder, which is responsible for segmenting the per-pixel visual embeddings $e_v^{pxl}$ into $C$ semantic clusters.


The whole framework of ViL-Seg is trained in an end-to-end manner, using the objective function as follows:
\begin{equation}
\mathcal{L} (\Phi_{v,t,c}) = \mathcal{L}_{vision} (\Phi_v) + \mathcal{L}_{cross} (\Phi_{v,t}) + \mathcal{L}_{cluster} (\Phi_c)
    \label{eq:overall_objective}
\end{equation}
which is composed of the vision-based contrastive learning  $\mathcal{L}_{vision}$ and the cross-modal contrastive alignment $\mathcal{L}_{cross}$ to enhance the fine-grained semantics and the high-level category information in the visual embeddings respectively; and an unsupervised clustering objective $\mathcal{L}_{cluster}$ optimized w.r.t. $\Phi_c$ to promote reasonable clustering results. Next, we will describe  each part in detail.



\subsection{Vision-based and Cross-modal Contrasting} 
\label{sec:vision-cross-contrasting}
As a dense classification task, semantic segmentation requires the learned visual embeddings to contain both \textit{fine-grained semantics} and \textit{high-level category information}. To this end, we have employed a vision-based contrasting and cross-modal contrasting to enhance the two properties of the visual representations respectively.

\textbf{Vision-based contrasting of global and local views:} Self-supervision with contrastive learning has shown promising results in representation learning~\cite{chen2020simple}. To meet the requirement of dense semantic understanding in segmentation, we devise a vision-based self-supervised learning strategy by contrasting local and global image patches to learn local to global semantic correspondence. 

Specifically, given an input visual image, we first transform it  into different distorted views or local patches, using the multi-crop strategy~\cite{caron2020unsupervised}, denoted as function $g(\cdot)$. This generates an image set of different views, which in our case contains one global view $x$ and $k$ local views $x^{local}=g(x)=[x^{l1}, x^{l2}, ..., x^{lk}]$ of low resolution. All these images are then fed into the visual encoder,  resulting in a global feature $e_v(x)$ of the global view $x$, and a local feature $e_v(x^{local})$ which is the concatenation of features of all local views $[e_v(x^{l1}), e_v(x^{l2}), \dots, e_v(x^{lk})]$. Considering that imposing the regularization directly onto the image features might be too strict to impede the convergence, we pass the global and local features to a projection function $\Phi_a$ before computing the loss function, which is composed of a linear projection layer and a softmax activation layer inspired by knowledge distillation~\cite{hinton2015distilling}. Our vision-based contrastive learning mechanism finally encourages the consistency of semantic information between the global and local features, aiming to encourage the model to capture the local to global correspondence and hence promote the fine-grained semantics of visual embeddings for the dense classification task. The objective function is expressed as:
\begin{equation}
\mathcal{L}_{vision} = H (\Phi_a(e_v (x)), \Phi_a(e_v(x^{local})))
\end{equation}
where $H(\cdot)$ denotes the cross-entropy loss.

\textbf{Cross-modal contrasting of natural language supervision:} Learning from natural language supervision has been demonstrated with effectiveness in large-scale vision-language pre-training tasks~\cite{li2020oscar,jia2021scaling,li2021unimo}. Our ViL-Seg inherits the cross-modal contrastive learning strategy, aiming to learn the visual embeddings  $e_v$ and text embeddings $e_t$ such that they can be close to each other if they are from the paired image and caption, and far away if not. 

Specifically, given a minibatch containing $b$ image-text pairs $\{x_j, t_j\}_{j=1}^{b}$, the image feature $e_v (x_m)$ and text feature $e_t (t_n)$ is a positive pair if $m=n$, and otherwise a negative pair. Then, the cross-modal contrastive alignment is performed over each positive pair in the minibatch as:
\begin{equation}
\small
\ell(x_m,\{t_n\}_{n=1}^b)=-log\frac{exp(e_v (x_m)\odot e_t (t_m) / \tau)}{\sum_{n=1}^{b} exp(e_v (x_m) \odot e_t (t_n)/ \tau)},
\end{equation}
where $\odot$ denotes the cosine similarity: $a \odot b = \frac{\langle a,b \rangle}{||a||_2||b||_2}$; $\tau$ denotes the temperature parameter. The final objective function $\mathcal{L}_{cross}$ is the average of $\ell$ over all positive pairs:
\begin{equation}
\small
\mathcal{L}_{cross}=\sum_{m=1}^{b}\frac{1}{b} \ell (x_m, \{t_n\}_{n=1}^b),
\label{eq:cross_contrastive}
\end{equation}
By aligning the 
visual and text embeddings as Eq.~\ref{eq:cross_contrastive}, the category information contained in the captions can be successfully transferred to the visual embeddings space, therefore allowing us to classify visual features by comparing their similarity with the word embeddings of different categories.


\subsection{Online Clustering of Visual Embeddings}
\label{sec:clustering}
Semantic segmentation requires assigning a label to each image pixel. However, the cross-modal alignment above can only provide classification ability over the global visual feature $e_v$, instead of per-pixel embeddings $e_v^{pxl}$. To address this problem, we propose to cluster the per-pixel visual features into distinct groups according to their semantics. Then, the features of each semantic region can be respectively abstracted as a region-level feature for cross-modal alignment to fulfill the dense classification pipeline. 

Specifically, we employ an online clustering strategy to efficiently separate the visual embeddings by maximizing the mutual information across cluster assignments. Given the per-pixel visual embeddings  $e_v^{pxl} \in \mathbb{R} ^ {HW \cdot D}$, we aim to cluster these features into clustering space $Y = \{1, 2, \dots, C\}$. To this end, we construct a clustering head $\Phi_c$ over the image encoder, which is composed of a convolution layer with $C$ channel followed by a softmax function. Denote $q, q' \in \mathbb{R} ^ {1\cdot D}$ as a pair of pixel embeddings from $e_v^{pxl}$ which contain the same semantic, the goal of our clustering head is to preserve what is common between $q$ and $q'$ while removing their instance-specific information, which is equivalent to maximizing their mutual information as:
\begin{equation}
    \text{max}_{\Phi_c} ~~ I (\Phi_c(q), \Phi_c(q'))
    \label{eq:pixel_mutual_information}
\end{equation}
In our case, the paired embeddings $(q, q')$ are unavailable since the category of each image pixel is unknown. Therefore, we adopt generated embedding pairs to compute the clustering objective, by extracting the embeddings for the input image $x$ and its transformed image $g(x)$ respectively, obtaining $e_v^{pxl} (x)$ and $e_v^{pxl} (g(x))$. It is worthy to mention that g($\cdot$) here do not adopt the multi-crop strategy, but the random additive and multiplicative colour transformations with horizontal flipping, which are all affine transformations. Since g($\cdot$) contains geometric transformation, the embedding $e_v^{pxl} (x)_i$ at pixel $i$ will correspond to $g^{-1}(e_v^{pxl} (g(x)))_{i}$. This is because translating the input image will also change the geometric order of the output feature. We need to undo the geometric function by applying $g^{-1}(\cdot)$ over the feature of transformed image such that it could be paired with $e_v^{pxl}(x)$ pixel-by-pixel. Please note that the reason we
compute clustering loss between pixels of different views
instead of pixels of the same class is that the class information of pixels are unknown in our case,since no dense
annotations are provided. Besides, maximizing the common information between transformed views is an effective strategy to promote clustering samples of the same class, as demonstrated in unsupervised learning~\cite{chen2020simple}, which meets our goal to perform semantic segmentation task without dense annotations.


We now describe how to compute the mutual information of Eq.~\ref{eq:pixel_mutual_information}. For simplicity of description, we denote $(q_i, q_i')$ as a pair of embeddings at  pixel $i$ of $e_v^{pxl} (x)$ and $g^{-1}(e_v^{pxl} (g(x)))$. Since our clustering head outputs soft label distributions using softmax activation function, the mutual information between $q_i$ and $q_i'$ (i.e., the probability of predicting $q_i$ from $q_i'$ and vice versa) is given by their joint probability distribution $J_{i} \in [0,1]^{C \times C}$:
\begin{equation}
I(\Phi_c{(q_i)}, \Phi_c{(q_i')}) = I(J_{i}), J_{i} = \Phi_c(q_i) \cdot \Phi_c (q_i')^T
\end{equation}
where $J_{i}^{cc'} = P(\Phi_c (q_i) = c,\Phi_c (q_i') = c' )$. In each minibatch, the joint probability distributions $J$ is computed as:
\begin{equation}
    J = \frac{1}{BHWD}\sum _ {i=1} ^ { BHWD} 
    \Phi_c(q_i) \cdot \Phi_c (q_i')^T
\end{equation}
Finally, the clustering objective is equivalent to maximizing the mutual information~\cite{tschannen2019mutual} of matrix $J$, and is extended to:
\begin{equation}
     \mathcal{L}_{cluster} = \text{max}~~I(J) = \text{max} ~~ \sum_{c=1}^C \sum_{c'=1}^C J^{cc'} \cdot\text{ln} \frac{J^{cc'}}{J^c \cdot J^{c'}}
    \label{eq:batch_mutual_information_maximization}
\end{equation} 
where $J^c=P(\Phi_c(q_i)=c)$ and $J^{c'}=P(\Phi_c(q_i')=c')$ are computed by summing over the $c$-th row and $c'$-th column of the matrix $J$ respectively. 

We take the relation between mutual information and entropy~\cite{paninski2003estimation} to explain why maximizing the mutual information can promote reasonable clustering results. Given $I (\Phi_c(q_i),\Phi_c(q_i')) =E(\Phi_c(q_i)) - E(\Phi_c(q_i)|\Phi_c(q_i')) $, maximizing the mutual information is equivalent to maximizing the individual clustering results entropy $E(\Phi_c(q_i))$ while minimizing the conditional clustering  results entropy $E(\Phi_c(q_i)|\Phi_c(q_i')) $. The smallest value of the latter is attained when  $E(\Phi_c(q_i)|\Phi_c(q_i'))=0$, i.e., the cluster assignments for $q_i$ and $q_i'$ are predictable for each other. Therefore, it encourages embeddings with similar semantics to be assigned to the  same cluster. Furthermore, the largest value of the $E(\Phi_c(q_i))$ is attained when all clusters are assigned in equal possibility among all embeddings in the whole dataset, hence it may avoid the degenerated solution that all features are assigned to the same cluster.

\textbf{Inference pipeline:} During inference, the segmentation for an input image $x$ can be produced by feeding it to the image encoder to extract per-pixel visual embeddings $e_v^{pxl}(x)$, which are then passed to the clustering head to obtain the clustering mask $M \in \{0,1\}^{H\times W\times C}$ with $C$ clusters using argmax function. According to the semantic region indicated by each cluster $M_c \in \{0,1\}^{H\times W}$, we can extract its region-level feature $e_v^{rgn} (M_c)$ by filtering and averaging the per-pixel visual embeddings in pixel indexes where $M_c = 1$ (cf. the region-level averaging pooling in Fig.~\ref{fig:framework}), i.e., $e_v^{rgn} (M_c)$ = $\frac{\sum e_v^{pxl}(x) \cdot M_c
}{\sum M_c}$. Finally, the category name of each region $M_c$ is given by comparing its region-level feature $e_v^{rgn} (M_c)$ with the word embddings of different classes, using prompt ``a photo of a {category}" as CLIP~\cite{radford2021learning}. 
\section{Experiments}

\subsection{Experimental Setup}
\textbf{Dataset and evaluation protocol:}
\label{sec:Dataset and Evaluation Protocols}
Following the literature of zero-shot segmentation~\cite{cheng2021sign,xian2019semantic,gu2020context}, we conduct experiments on three datasets, including PASCAL VOC~\cite{everingham2010pascal}, PASCAL Context~\cite{mottaghi2014role}, and COCO Stuff~\cite{caesar2018coco}. For PASCAL VOC and PASCAL Context datasets, we evaluate our method on their validation set containing 1449 images and 5105 images respectively. For COCO Stuff datasets, we adopt the setting in ~\cite{cheng2021sign} to use 5000 images for testing.

Since there is not a standard evaluation protocol for our open-world semantic segmentation task without using any dense annotations, we follow the zero-shot segmentation settings defined in ~\cite{xian2019semantic,gu2020context} to compare the segmentation performance on the unseen classes of the three datasets. Specifically, the unseen classes contain: 5 classes (potted plant, sheep, sofa, train, tv-monitor) out of the 20 object categories in PASCAL VOC; 4 classes (cow, motorbike, sofa, cat) out of the 59 object categories  in PASCAL Context; and 15 classes (frisbee, skateboard, cardboard, carrot, scissors, suitcase, giraffe, cow, road, wall concrete, tree, grass, river, clouds, playingfield) out of the 183 object categories in COCO Stuff dataset. We adopt the standard  metrics including mean intersection-over-union (mIoU) \cite{long2015fully} and pixel accuracy (pix. acc.) to evaluate the segmentation results.

\begin{table*}[t!]
\renewcommand\arraystretch{1.0}
\centering
\caption{Comparison of unseen-class segmentation results with zero-shot segmentation methods on Pascal VOC, Pascal Context and COCO Stuff datasets. "ST" stand for self-training.}
\vspace{-1mm}
\label{tab:comparison}
\scalebox{0.85}{
\begin{tabular}{l|cc|cc|cc} 
\hline
\hline
& \multicolumn{2}{c|}{PASCAL VOC} & \multicolumn{2}{c|}{PASCAL Context}   & \multicolumn{2}{c}{COCO Stuff} \\ 
\hline
Method      & mIoU [\%] & {pix. acc. [\%]}  & mIoU [\%] & {pix. acc. [\%]}& mIoU [\%] & {pix. acc. [\%]}  \\ 
\hline

SPNet~\cite{semanticxian}     & 15.63 &- & 4.00 &- & 8.73     & -\\
ZS3~\cite{bucher2019zero}      & 17.65 &21.47 & 7.68 &19.22 & 9.53       &22.75         \\
CaGNet (pi)~\cite{gu2020context}     & 26.59 &42.97 & 14.42 &39.76  & 12.23   &25.45        \\
CaGNet (pa)~\cite{gu2020context} & 29.90 & 51.76  &14.98 &39.81 &13.89 &29.62\\
SIGN~\cite{cheng2021sign}     & 28.86 &- & 14.93 &- & 15.47   &-            \\
CLIP + Seg    &    27.40  &  48.35   &   14.52   &   37.48    &  13.20   &  28.75    \\
 \hline
ViL-Seg (Ours)     & \textbf{34.42$_{(+5.56)}$} &  \textbf{76.03$_{(+24.27)}$}   & \textbf{16.32}$_{(+1.39)}$ &  \textbf{45.64$_{(+5.83)}$} &  \textbf{16.43$_{(+0.96)}$}&  \textbf{32.58$_{(+2.96)}$} \\

\hline
\hline
ZS3 + ST & 21.15 & - & 9.53 & - & 10.55 & - \\
CaGNet + ST & 30.31 & -  & 16.30& - & 13.40 & - \\
SIGN + ST & 33.12 & - & 16.71   & - & 15.15 &  -\\
ViL-Seg + ST & \textbf{37.30$_{(+4.18)}$} & 85.62 & \textbf{18.94$_{(+2.13)}$ } & 50.14 &\textbf{18.05$_{(+2.90)}$} & 35.23\\
\hline
\hline

\end{tabular}
}
\end{table*}

\textbf{Implementation detail:}
We adopt the transformer architecture (ViT-B/16) for the image encoder and text encoder, following the popular vision-language learning framework~\cite{radford2021learning}, with the embedding dimension of 512.
The cluster number $C$ in the online clustering head is set as 25, and we shall study this hyper-parameter in detail in the ablation analysis. In vision-based contrasting, we crop 6 local patches with the multi-crop strategy, and the output dimension of the projection layer is 2048.
We train the model with Adam~\cite{loshchilov2018fixing} optimizer, using learning rate of 5e-4, weight decay coefficient of 0.04, and warm-up iterations of 4000. The ViL-Seg model is trained with no other data but CC12M dataset~\cite{changpinyo2021conceptual}, which contains about 12 million image-caption pairs collected from the Internet. The whole framework is trained using 48 Tesla V100 16GB with batch size 768.


\subsection{Comparison with Other Methods}
\textbf{Experimental setting:} Due to the lack of previous study that purely utilizes web-based image-caption data to learn to segment novel object categories, we compare our method with several popular zero-shot segmentation (ZSS) methods, which also segment new object categories but via exploiting the relationships between the word embeddings of seen base classes and unseen class. Specifically, the comparison methods include (1) SPNet~\cite{semanticxian}, a semantic projection network which maps each image pixel to a semantic word embedding space for ZSS; (2) ZS3~\cite{bucher2019zero}, which addresses unseen-class segmentation by combining a segmentation model with an approach to generate visual representations from semantic word embedding; (3) CaGNet~\cite{gu2020context}, which devises a contextual module into the segmentation network to capture more diverse contextual information from semantic word embedding; and (4) SIGN~\cite{cheng2021sign}, a very latest ZSS method which incorporates spatial information into semantic features using positional encodings to improve the segmentation of unseen classes. (5)
 CLIP~\cite{radford2021learning} + Seg, we simply use the CLIP's image encoder(ViT-B/16) with its global attention pooling layer removed, to serve as a backbone for semantic segmentation. Classification  for dense prediction can be directly obtained from the text embeddings of CLIP's text encoder. 
All these methods follow the same zero-shot segmentation setting described in Sec.~\ref{sec:Dataset and Evaluation Protocols}, and for a fair comparison, we compare the performance of all these methods under both scenarios of using or without using self-training as followup. For each comparison method, the results are either referenced from their official paper or the number reproduced by other previous works.

\begin{figure*}[t]
		\begin{center}
            \includegraphics[ width=0.95\linewidth]{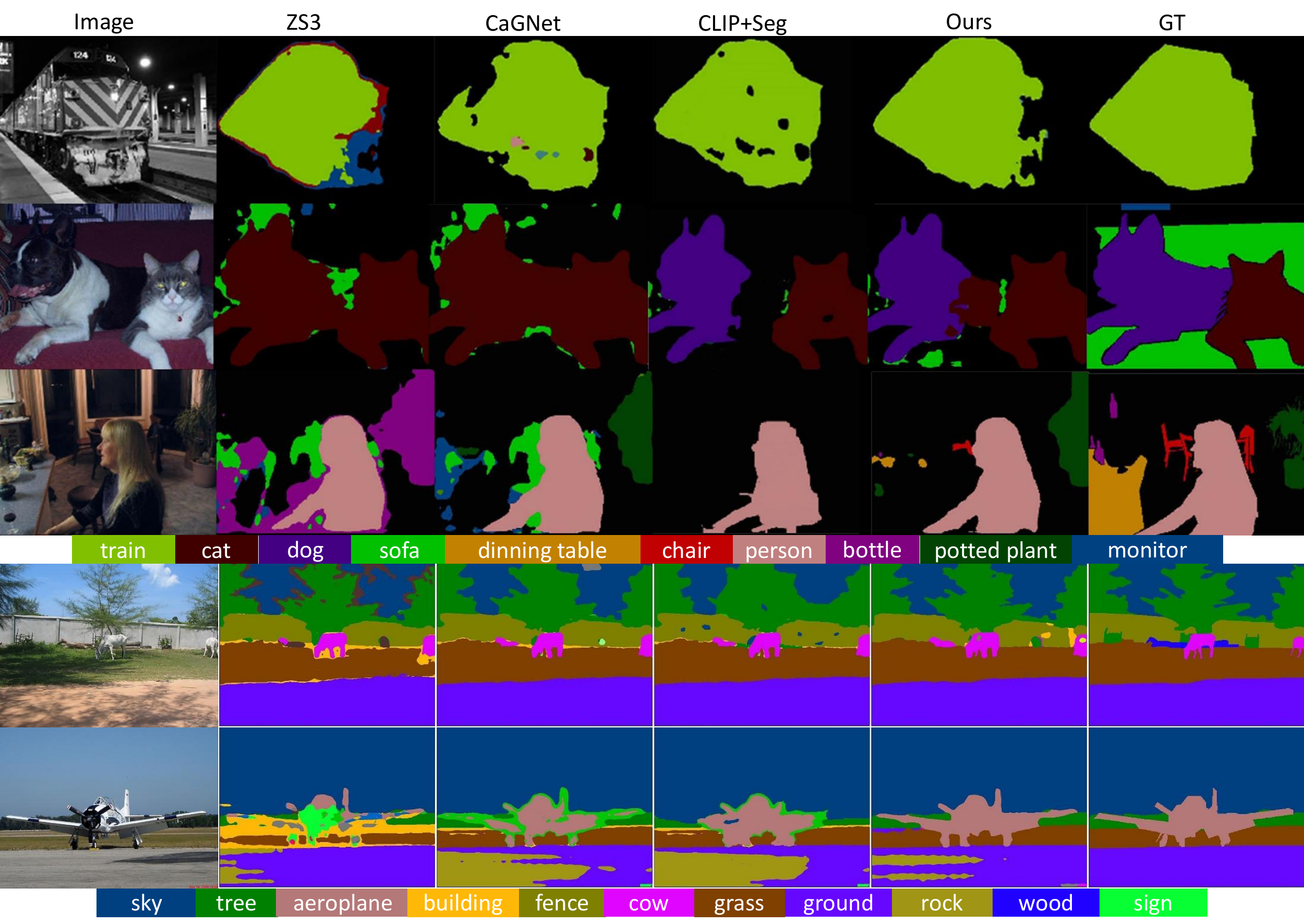}
		\end{center}
		\vspace{-4mm}
		\caption{Qualitative comparison with baseline and other methods. The top three samples are from PASCAL VOC and the bottom two samples are from PASCAL Context. }
		\label{fig:method_compare}
\end{figure*}

\textbf{Comparison results:} Table~\ref{tab:comparison} presents the comparison results of these methods on PASCAL VOC~\cite{everingham2010pascal}, PASCAL Context~\cite{mottaghi2014role} and COCO stuff~\cite{caesar2018coco} dataset (``-" denote the result was not reported in their paper). From this table, we may draw the following observations: (1) Our ViL-Seg outperforms these zero-shot segmentation methods on all three datasets in terms of both mIoU and pixel accuracy. This confirms the feasibility to exploit the naturally-existing image-caption pairs from the Internet to learn the segmentation model that can segment various open-world object categories. It is notable that these ZSS methods need to be trained on the densely-annotated training sets containing diverse base categories, but our ViL-Seg does not use any data with dense annotations for training. (2) ViL-Seg shows a larger increase on PASCAL VOC over other methods compared with the other two datasets. A plausible reason is that PASCAL VOC only contains 15 seen bases classes for these ZSS methods to train the model, which is relatively less than the 55 and 168 seen classes in PASCAL Context and COCO Stuff. In such case, our larger improvements in PASCAL VOC may reflect the limitation of  those ZSS methods that require a wide range of base categories with dense annotations to attain a good performance,  and further confirms the advantage of ViL-Seg that requires no data labeling. Fig. ~\ref{fig:method_compare} shows a qualitative comparison between ViL-Seg
and baselines (SIGN~\cite{cheng2021sign} does not release its code). We can see that ViL-Seg achieves high accuracy.

\subsection{Ablation Analysis of ViL-Seg}
We conduct ablation studies on the three datasets to investigate several key questions of ViL-Seg: \textbf{1)} the importance of the vision-based contrastive learning in ViL-Seg; \textbf{2)} the benefit of the online clustering head compared with offline clustering method like K-means; \textbf{3)} the choice and effect of cluster number in the online clustering head; \textbf{4)} the performance of ViL-Seg on different unseen classes. In ablation analysis, all object categories in the three datasets are considered as unseen classes. The performance on each dataset is the average over all its contained classes. 

\begin{figure*}[t]
		\begin{center}
            \includegraphics[ width=0.95\linewidth]{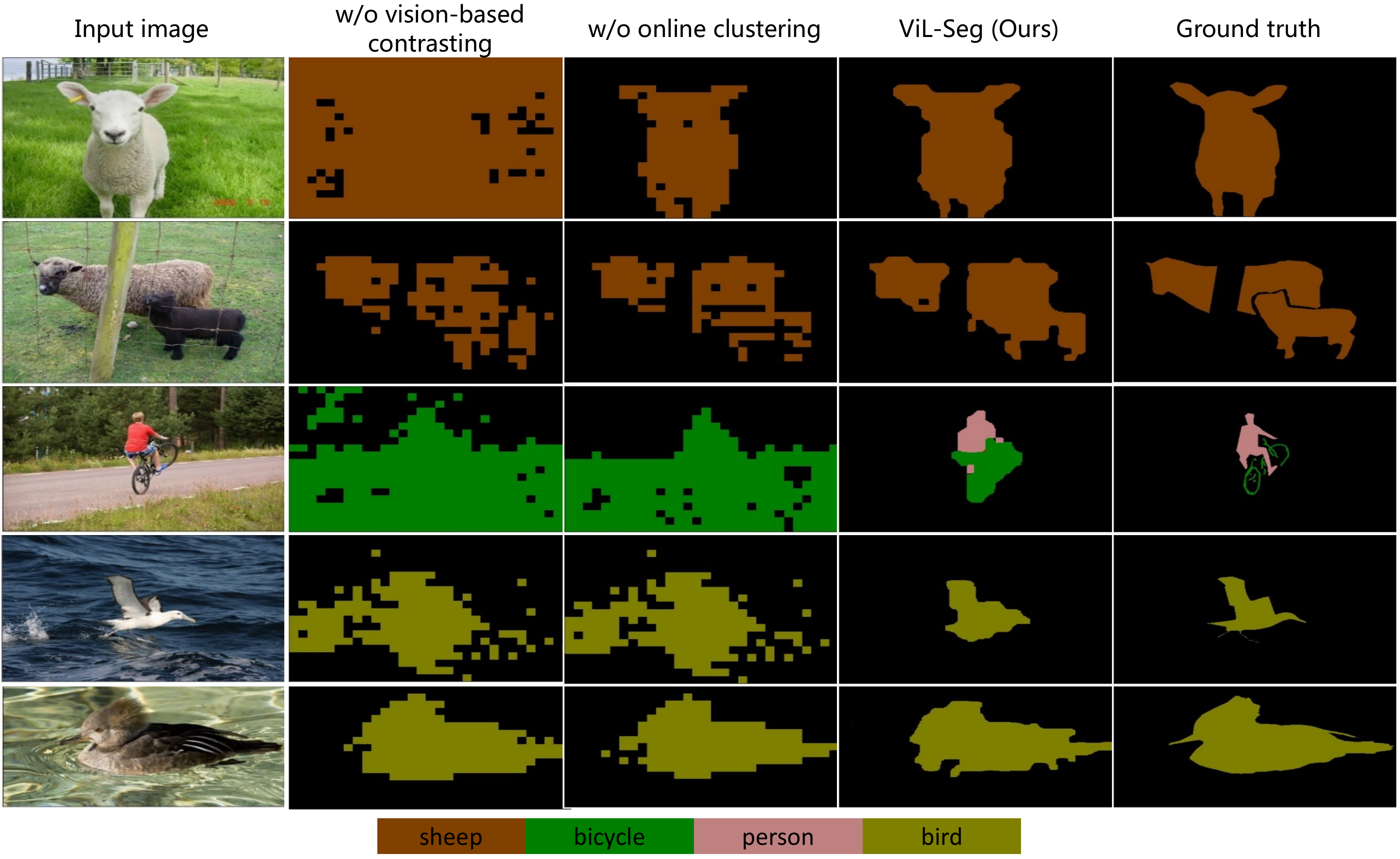}
		\end{center}
		\vspace{-4mm}
		\caption{Qualitative comparison among ViL-Seg, ViL-Seg without online clustering, and ViL-Seg without vision-based contrasting, with  samples from PASCAL VOC dataset.  }
		\vspace{-3mm}
		\label{fig:qualitative_segmentation}
\end{figure*}

\textbf{Importance of vision-based contrasting:} 
Apart from the cross-modal contrasting to align the visual and text embedding space, the image encoder in our framework is further supervised with a self-supervision signal by contrasting local and global image patches. From the qualitative segmentation results in Fig.~\ref{fig:qualitative_segmentation}, we can clearly see that without using this vision-based contrasting (second column), the clustering results  cannot accurately separate the semantic object from the background region. Besides, the quantitative results in Table~\ref{tab:ablation_all}  show that removing this supervision (ViL-Seg w/o $\mathcal{L}_{vision}$) will lead to large performance decreases on all three datasets. These results reflect that the cross-modal contrasting can only guarantee the semantics of global image feature which is insufficient for the dense classification problem, while 
our additional self-supervision signal with vision-based contrasting is crucial to promote the fine-grained semantics in the visual embeddings.


\begin{table}[t!]
\label{tab:ab1}
\centering
\vspace{-1mm}
\caption{Ablation analysis of the vision-based contrastive learning (i.e., $\mathcal{L}_{vision}$), and online clustering design on the three datasets.}
\label{tab:ablation_all}
\scalebox{0.9}{
\begin{tabular}{c|c|c ccc} 
\hline
\hline
\multicolumn{2}{c}{}     & ViL-Seg w/o $\mathcal{L}_{vision}$ &Offline (K-means) & ViL-Seg \\
\hline
\multicolumn{2}{c|}{Params}& -   &   86.19M &86.27M \\
\multicolumn{2}{c|}{Speed (case / s)}&   - &8.5 &9.8 \\
\hline
  PASCAL & mIoU [\%]        &22.05           &30.97 &33.61 \\
  VOC&  pix. acc. [\%]      & 50.76          &   69.88 &  75.97          \\
  \hline
    PASCAL & mIoU [\%]      & 13.14         &    14.82 & 15.89                 \\
  Context& pix. acc. [\%]   & 38.90           &      41.64 &  43.54            \\
  \hline
    COCO  & mIoU [\%]       & 13.52         &  15.81  &     16.41                  \\
  Stuff& pix. acc. [\%]     & 28.07         &  30.45  &   31.20                  \\
\hline
\hline
\end{tabular}
}
\vspace{-2mm}
\end{table}

\textbf{Online clustering $v.s.$ offline clustering:} Traditionally, the usual way to segment a group of features into distinct clusters is the offline methods like K-means~\cite{kodinariya2013review}. Table~\ref{tab:ablation_all} compares our online clustering design with traditional offline method, by replacing our online clustering head with K-means to cluster the per-pixel visual embeddings. We may draw three observations: (1) Our online clustering design attains higher segmentation performance than the offline method on all three datasets. We consider that the online clustering head is tightly-coupled with the visual encoder and can learn to improve the quality of visual embeddings as the training goes on, which is what the offline methods cannot attain. The qualitative results in Fig.~\ref{fig:qualitative_segmentation} can also reflect that our online method (the fourth column) can better refine the learned visual embeddings and produce more smooth segmentation masks than the offline method (the third column). (2) The framework with our online clustering design also achieves a higher inference speed than the offline K-means method (8.5 $v.s.$ 9.8 cases / s). This is because K-means needs to be performed offline as  post-processing on the network  features, which would limit the inference efficiency. In contrast, our online clustering design makes the training and inference of our method end-to-end and allows us to adaptively cluster the visual  embeddings for each sample. (3) Additionally, compared with the offline method, our online clustering design only increases 0.08M parameters to the model, which is less than 0.1\% of the number of original network parameters.
\begin{figure}[t]
\centering
    \begin{minipage}[t]{0.45\linewidth}
    \centering
    \includegraphics[ width=1.0\linewidth]{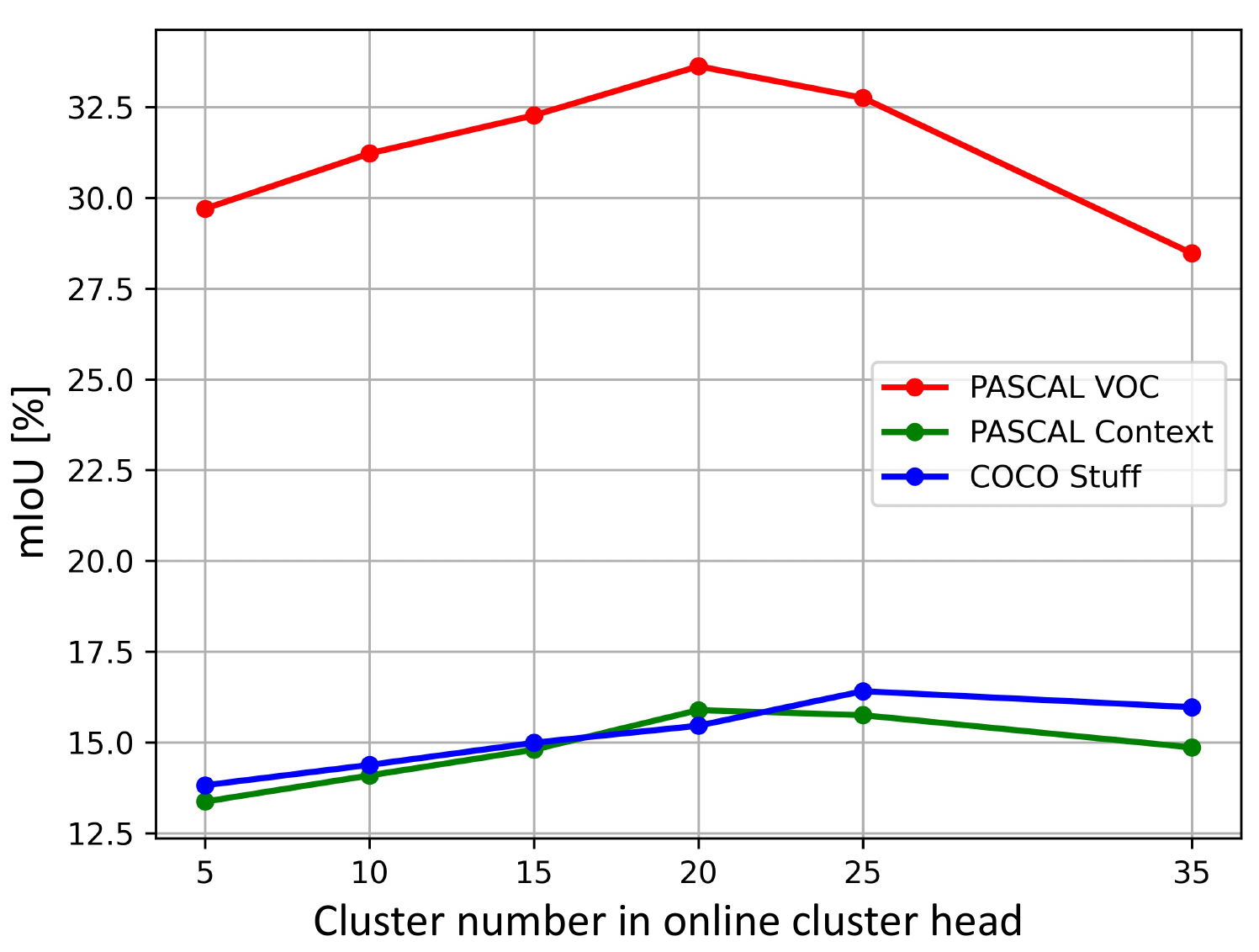}
	\vspace{-3mm}
	\caption{Segmentation performance of  ViL-Seg under different choices of cluster number $C$ in the online clustering head, on PASCAL VOC, PASCAL Context and COCO stuff datasets.}
	\vspace{-2mm}
	\label{fig:ablation_on_cluster_number}
	\end{minipage}
	\hspace{5mm}
	\begin{minipage}[t]{0.45\linewidth}
	\centering
    \includegraphics[ width=1.0\linewidth]{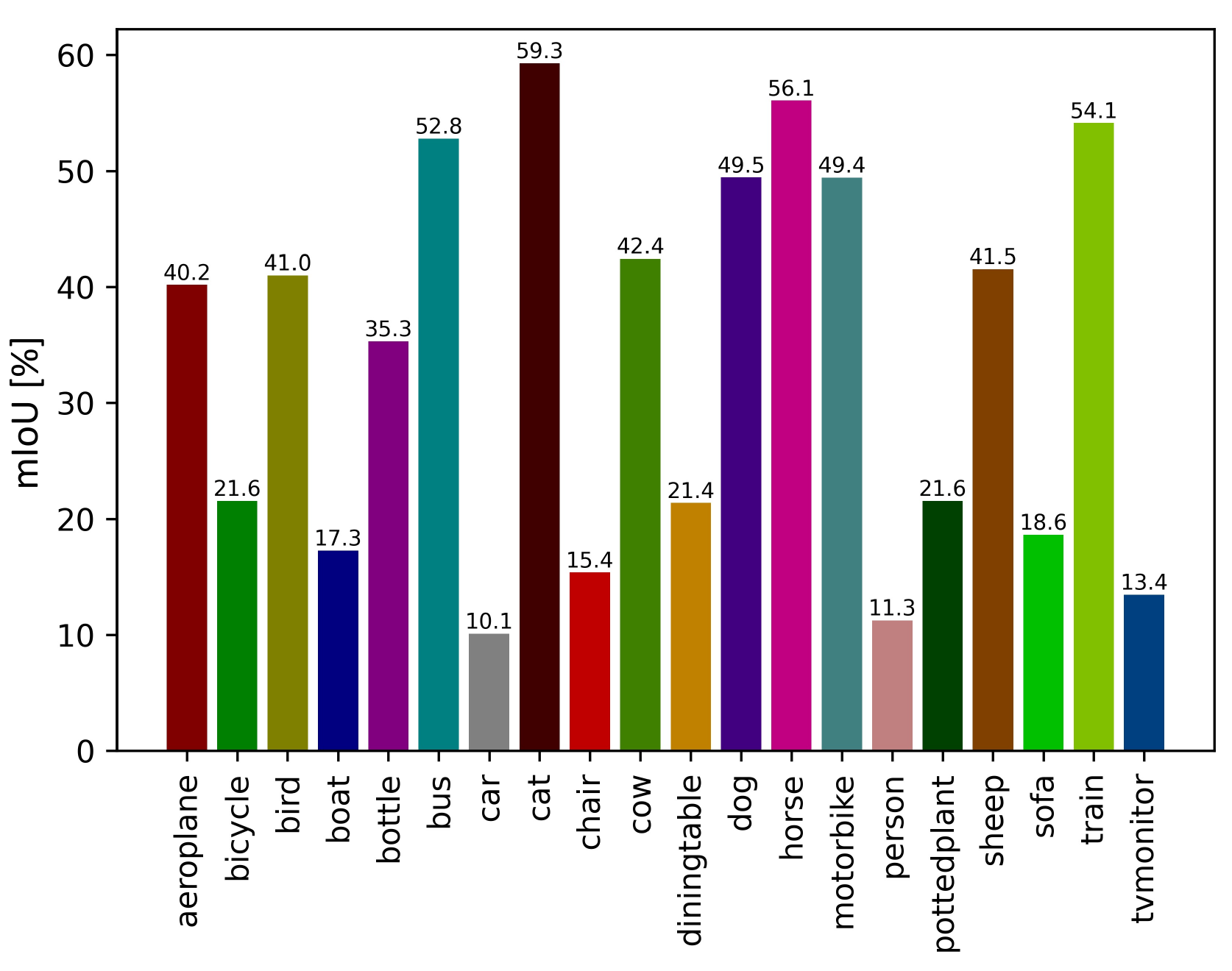}
	\vspace{-3mm}
	\caption{Segmentation performance of ViL-Seg on all 20 unseen classes of PASCAL VOC dataset. It is noticed that ViL-Seg can attain mIoU larger than 20\% on 14 out of 20 unseen classe.}
	\vspace{-2mm}
	\label{fig:performance_of_each_class}
	\end{minipage}
\end{figure}


\textbf{Effect of cluster number in online clustering head:}
The cluster number $C$ is important in our method and affects the results of the online clustering head. 
Intuitively, fewer clusters might be incapable to cover the diverse semantics in the web-based image-caption data,  while too many clusters might increase the learning difficulty given that the clustering head is only learned with an unsupervised objective of mutual information maximization. To validate the above intuitions and investigate the suitable choice of $C$, we repeated the experiment of ViL-Seg by varying $C$$\in$$\{5, 10, 15, 20, 25, 35\}$. As shown in Fig.~\ref{fig:ablation_on_cluster_number}, the model with middle-level of cluster number ($C$$=$$\{20, 25\}$) performs 
better than the model with smaller ($C$$=$$\{5, 10, 15\}$ or larger cluster number ($C$$=$$30$). These results confirm our analysis above, and we finally adopt $C$$=$$20$ in our method.

\textbf{Performance on different unseen classes:}
In Fig.~\ref{fig:performance_of_each_class}, we show the mIoU of ViL-Seg on all 20 unseen classes of PASCAL VOC. It is observed that ViL-Seg can achieves more than 50\% mIoU for classes like ``bus", ``cat", ``horse" and ``train", and attain mIoU larger than 20\% on 14 out of 20 unseen classes. This owes to the diverse semantic information contained in the web-based data, which allows ViL-Seg to well segment these object categories even without using any of their training data with dense annotations. We also notice that the performance is relatively low in class like ``person" or ``car". This is probably caused by the imbalanced recognition capacity of vision-language models, which was also reported in previous studies~\cite{radford2021learning}. For example, the image captions might usually use words like ``man", ``woman" to denote a person; and use the word of a brand name to denote a car, making the model less sensitive to these object categories. We may consider ensembling the results of different synonyms for an object category to alleviate this issue~\cite{lin2021m6}.

\section{Conclusion}
We have made the first attempt to learn to segment open-world object categories by purely leveraging the image-caption data from the Internet, without using any data with dense annotations. The proposed ViL-Seg attains the segmentation ability by employing two complementary contrastive learning strategies to promote the quality of visual embeddings, with an online clustering head to dynamically segment them into distinct semantic groups. Owing to the tremendous data resources on the Internet, our solution has outperformed zero-shot segmentation methods to segment the diverse semantic concepts in reality on three benchmark datasets, also opened a door for semantic segmentation task to reduce the human labeling to the greatest extent.

\noindent \textbf{Acknowledgements} We gratefully acknowledge the support of MindSpore\footnote{\url{https://www.mindspore.cn/}}, CANN~(Compute Architecture for Neural Networks) and Ascend AI Processor used for this research.

\clearpage
{\small
\bibliographystyle{ieee_fullname}
\bibliography{egbib}

\begin{thebibliography}{10}\itemsep=-1pt

\bibitem{bucher2019zero}
Maxime Bucher, Tuan-Hung Vu, Matthieu Cord, and Patrick P{\'e}rez.
\newblock Zero-shot semantic segmentation.
\newblock {\em Advances in Neural Information Processing Systems}, 32:468--479,
  2019.

\bibitem{caesar2018coco}
Holger Caesar, Jasper Uijlings, and Vittorio Ferrari.
\newblock Coco-stuff: Thing and stuff classes in context.
\newblock In {\em Proceedings of the IEEE conference on computer vision and
  pattern recognition}, pages 1209--1218, 2018.

\bibitem{caron2020unsupervised}
Mathilde Caron, Ishan Misra, Julien Mairal, Priya Goyal, Piotr Bojanowski, and
  Armand Joulin.
\newblock Unsupervised learning of visual features by contrasting cluster
  assignments.
\newblock {\em arXiv preprint arXiv:2006.09882}, 2020.

\bibitem{changpinyo2021conceptual}
Soravit Changpinyo, Piyush Sharma, Nan Ding, and Radu Soricut.
\newblock Conceptual 12m: Pushing web-scale image-text pre-training to
  recognize long-tail visual concepts.
\newblock In {\em Proceedings of the IEEE/CVF Conference on Computer Vision and
  Pattern Recognition}, pages 3558--3568, 2021.

\bibitem{chen2017deeplab}
Liang-Chieh Chen, George Papandreou, Iasonas Kokkinos, Kevin Murphy, and Alan~L
  Yuille.
\newblock Deeplab: Semantic image segmentation with deep convolutional nets,
  atrous convolution, and fully connected crfs.
\newblock {\em IEEE transactions on pattern analysis and machine intelligence},
  40(4):834--848, 2017.

\bibitem{chen2020simple}
Ting Chen, Simon Kornblith, Mohammad Norouzi, and Geoffrey Hinton.
\newblock A simple framework for contrastive learning of visual
  representations.
\newblock In {\em International conference on machine learning}, pages
  1597--1607. PMLR, 2020.

\bibitem{neilchen2013}
Xinlei Chen, Abhinav Shrivastava, and Abhinav Gupta.
\newblock Neil: Extracting visual knowledge from web data.
\newblock In {\em 2013 IEEE International Conference on Computer Vision}, pages
  1409--1416, 2013.

\bibitem{chen2020uniter}
Yen-Chun Chen, Linjie Li, Licheng Yu, Ahmed El~Kholy, Faisal Ahmed, Zhe Gan, Yu
  Cheng, and Jingjing Liu.
\newblock Uniter: Universal image-text representation learning.
\newblock In {\em European conference on computer vision}, pages 104--120.
  Springer, 2020.

\bibitem{cheng2021sign}
Jiaxin Cheng, Soumyaroop Nandi, Prem Natarajan, and Wael Abd-Almageed.
\newblock Sign: Spatial-information incorporated generative network for
  generalized zero-shot semantic segmentation, 2021.

\bibitem{everingham2010pascal}
Mark Everingham, Luc Van~Gool, Christopher~KI Williams, John Winn, and Andrew
  Zisserman.
\newblock The pascal visual object classes (voc) challenge.
\newblock {\em International journal of computer vision}, 88(2):303--338, 2010.

\bibitem{geng2020recent}
Chuanxing Geng, Sheng-jun Huang, and Songcan Chen.
\newblock Recent advances in open set recognition: A survey.
\newblock {\em IEEE transactions on pattern analysis and machine intelligence},
  2020.

\bibitem{gu2021openvocabulary}
Xiuye Gu, Tsung-Yi Lin, Weicheng Kuo, and Yin Cui.
\newblock Open-vocabulary object detection via vision and language knowledge
  distillation, 2021.

\bibitem{gu2020context}
Zhangxuan Gu, Siyuan Zhou, Li Niu, Zihan Zhao, and Liqing Zhang.
\newblock Context-aware feature generation for zero-shot semantic segmentation.
\newblock In {\em Proceedings of the 28th ACM International Conference on
  Multimedia}, pages 1921--1929, 2020.

\bibitem{gupta2019lvis}
Agrim Gupta, Piotr Dollar, and Ross Girshick.
\newblock Lvis: A dataset for large vocabulary instance segmentation.
\newblock In {\em Proceedings of the IEEE/CVF Conference on Computer Vision and
  Pattern Recognition}, pages 5356--5364, 2019.

\bibitem{hinton2015distilling}
Geoffrey Hinton, Oriol Vinyals, and Jeff Dean.
\newblock Distilling the knowledge in a neural network.
\newblock {\em arXiv preprint arXiv:1503.02531}, 2015.

\bibitem{huang2021seeing}
Zhicheng Huang, Zhaoyang Zeng, Yupan Huang, Bei Liu, Dongmei Fu, and Jianlong
  Fu.
\newblock Seeing out of the box: End-to-end pre-training for vision-language
  representation learning.
\newblock In {\em Proceedings of the IEEE/CVF Conference on Computer Vision and
  Pattern Recognition}, pages 12976--12985, 2021.

\bibitem{huo2021wenlan}
Yuqi Huo, Manli Zhang, Guangzhen Liu, Haoyu Lu, Yizhao Gao, Guoxing Yang,
  Jingyuan Wen, Heng Zhang, Baogui Xu, Weihao Zheng, et~al.
\newblock Wenlan: Bridging vision and language by large-scale multi-modal
  pre-training.
\newblock {\em arXiv preprint arXiv:2103.06561}, 2021.

\bibitem{hwang2019segsort}
Jyh-Jing Hwang, Stella~X Yu, Jianbo Shi, Maxwell~D Collins, Tien-Ju Yang, Xiao
  Zhang, and Liang-Chieh Chen.
\newblock Segsort: Segmentation by discriminative sorting of segments.
\newblock In {\em Proceedings of the IEEE/CVF International Conference on
  Computer Vision}, pages 7334--7344, 2019.

\bibitem{jia2021scaling}
Chao Jia, Yinfei Yang, Ye Xia, Yi-Ting Chen, Zarana Parekh, Hieu Pham, Quoc~V
  Le, Yunhsuan Sung, Zhen Li, and Tom Duerig.
\newblock Scaling up visual and vision-language representation learning with
  noisy text supervision.
\newblock {\em arXiv preprint arXiv:2102.05918}, 2021.

\bibitem{glenn_jocher_2020_4154370}
Glenn Jocher.
\newblock {ultralytics/yolov5: v3.1 - Bug Fixes and Performance Improvements}.
\newblock \url{https://github.com/ultralytics/yolov5}, Oct. 2020.

\bibitem{kato2019zero}
Naoki Kato, Toshihiko Yamasaki, and Kiyoharu Aizawa.
\newblock Zero-shot semantic segmentation via variational mapping.
\newblock In {\em Proceedings of the IEEE/CVF International Conference on
  Computer Vision Workshops}, pages 0--0, 2019.

\bibitem{kodinariya2013review}
Trupti~M Kodinariya and Prashant~R Makwana.
\newblock Review on determining number of cluster in k-means clustering.
\newblock {\em International Journal}, 1(6):90--95, 2013.

\bibitem{li2020consistent}
Peike Li, Yunchao Wei, and Yi Yang.
\newblock Consistent structural relation learning for zero-shot segmentation.
\newblock {\em Advances in Neural Information Processing Systems}, 33, 2020.

\bibitem{li2021unimo}
Wei Li, Can Gao, Guocheng Niu, Xinyan Xiao, Hao Liu, Jiachen Liu, Hua Wu, and
  Haifeng Wang.
\newblock Unimo: Towards unified-modal understanding and generation via
  cross-modal contrastive learning, 2021.

\bibitem{li2020oscar}
Xiujun Li, Xi Yin, Chunyuan Li, Pengchuan Zhang, Xiaowei Hu, Lei Zhang, Lijuan
  Wang, Houdong Hu, Li Dong, Furu Wei, et~al.
\newblock Oscar: Object-semantics aligned pre-training for vision-language
  tasks.
\newblock In {\em European Conference on Computer Vision}, pages 121--137.
  Springer, 2020.

\bibitem{lin2017refinenet}
Guosheng Lin, Anton Milan, Chunhua Shen, and Ian Reid.
\newblock Refinenet: Multi-path refinement networks for high-resolution
  semantic segmentation.
\newblock In {\em Proceedings of the IEEE conference on computer vision and
  pattern recognition}, pages 1925--1934, 2017.

\bibitem{lin2021m6}
Junyang Lin, Rui Men, An Yang, Chang Zhou, Ming Ding, Yichang Zhang, Peng Wang,
  Ang Wang, Le Jiang, Xianyan Jia, et~al.
\newblock M6: A chinese multimodal pretrainer.
\newblock {\em arXiv preprint arXiv:2103.00823}, 2021.

\bibitem{long2015fully}
Jonathan Long, Evan Shelhamer, and Trevor Darrell.
\newblock Fully convolutional networks for semantic segmentation.
\newblock In {\em Proceedings of the IEEE conference on computer vision and
  pattern recognition}, pages 3431--3440, 2015.

\bibitem{loshchilov2018fixing}
Ilya Loshchilov and Frank Hutter.
\newblock Fixing weight decay regularization in adam, 2018.

\bibitem{minaee2021image}
Shervin Minaee, Yuri~Y Boykov, Fatih Porikli, Antonio~J Plaza, Nasser
  Kehtarnavaz, and Demetri Terzopoulos.
\newblock Image segmentation using deep learning: A survey.
\newblock {\em IEEE Transactions on Pattern Analysis and Machine Intelligence},
  2021.

\bibitem{mottaghi2014role}
Roozbeh Mottaghi, Xianjie Chen, Xiaobai Liu, Nam-Gyu Cho, Seong-Whan Lee, Sanja
  Fidler, Raquel Urtasun, and Alan Yuille.
\newblock The role of context for object detection and semantic segmentation in
  the wild.
\newblock In {\em Proceedings of the IEEE conference on computer vision and
  pattern recognition}, pages 891--898, 2014.

\bibitem{oza2019c2ae}
Poojan Oza and Vishal~M Patel.
\newblock C2ae: Class conditioned auto-encoder for open-set recognition.
\newblock In {\em Proceedings of the IEEE/CVF Conference on Computer Vision and
  Pattern Recognition}, pages 2307--2316, 2019.

\bibitem{pakhomov2021segmentation}
Daniil Pakhomov, Sanchit Hira, Narayani Wagle, Kemar~E Green, and Nassir Navab.
\newblock Segmentation in style: Unsupervised semantic image segmentation with
  stylegan and clip.
\newblock {\em arXiv preprint arXiv:2107.12518}, 2021.

\bibitem{paninski2003estimation}
Liam Paninski.
\newblock Estimation of entropy and mutual information.
\newblock {\em Neural computation}, 15(6):1191--1253, 2003.

\bibitem{perera2020generative}
Pramuditha Perera, Vlad~I Morariu, Rajiv Jain, Varun Manjunatha, Curtis
  Wigington, Vicente Ordonez, and Vishal~M Patel.
\newblock Generative-discriminative feature representations for open-set
  recognition.
\newblock In {\em Proceedings of the IEEE/CVF Conference on Computer Vision and
  Pattern Recognition}, pages 11814--11823, 2020.

\bibitem{radford2021learning}
Alec Radford, Jong~Wook Kim, Chris Hallacy, Aditya Ramesh, Gabriel Goh,
  Sandhini Agarwal, Girish Sastry, Amanda Askell, Pamela Mishkin, Jack Clark,
  et~al.
\newblock Learning transferable visual models from natural language
  supervision.
\newblock {\em arXiv preprint arXiv:2103.00020}, 2021.

\bibitem{ronneberger2015u}
Olaf Ronneberger, Philipp Fischer, and Thomas Brox.
\newblock U-net: Convolutional networks for biomedical image segmentation.
\newblock In {\em International Conference on Medical image computing and
  computer-assisted intervention}, pages 234--241. Springer, 2015.

\bibitem{scheirer2012toward}
Walter~J Scheirer, Anderson de Rezende~Rocha, Archana Sapkota, and Terrance~E
  Boult.
\newblock Toward open set recognition.
\newblock {\em IEEE transactions on pattern analysis and machine intelligence},
  35(7):1757--1772, 2012.

\bibitem{su2019vl}
Weijie Su, Xizhou Zhu, Yue Cao, Bin Li, Lewei Lu, Furu Wei, and Jifeng Dai.
\newblock Vl-bert: Pre-training of generic visual-linguistic representations.
\newblock {\em arXiv preprint arXiv:1908.08530}, 2019.

\bibitem{thomee2016yfcc100m}
Bart Thomee, David~A Shamma, Gerald Friedland, Benjamin Elizalde, Karl Ni,
  Douglas Poland, Damian Borth, and Li-Jia Li.
\newblock Yfcc100m: The new data in multimedia research.
\newblock {\em Communications of the ACM}, 59(2):64--73, 2016.

\bibitem{tschannen2019mutual}
Michael Tschannen, Josip Djolonga, Paul~K Rubenstein, Sylvain Gelly, and Mario
  Lucic.
\newblock On mutual information maximization for representation learning.
\newblock {\em arXiv preprint arXiv:1907.13625}, 2019.

\bibitem{van2021unsupervised}
Wouter Van~Gansbeke, Simon Vandenhende, Stamatios Georgoulis, and Luc Van~Gool.
\newblock Unsupervised semantic segmentation by contrasting object mask
  proposals.
\newblock {\em arXiv preprint arXiv:2102.06191}, 2021.

\bibitem{wang2021simvlm}
Zirui Wang, Jiahui Yu, Adams~Wei Yu, Zihang Dai, Yulia Tsvetkov, and Yuan Cao.
\newblock Simvlm: Simple visual language model pretraining with weak
  supervision.
\newblock {\em arXiv preprint arXiv:2108.10904}, 2021.

\bibitem{semanticxian}
Yongqin Xian, Subhabrata Choudhury, Yang He, Bernt Schiele, and Zeynep Akata.
\newblock Semantic projection network for zero- and few-label semantic
  segmentation.
\newblock In {\em 2019 IEEE/CVF Conference on Computer Vision and Pattern
  Recognition (CVPR)}, pages 8248--8257, 2019.

\bibitem{xian2019semantic}
Yongqin Xian, Subhabrata Choudhury, Yang He, Bernt Schiele, and Zeynep Akata.
\newblock Semantic projection network for zero-and few-label semantic
  segmentation.
\newblock In {\em Proceedings of the IEEE/CVF Conference on Computer Vision and
  Pattern Recognition}, pages 8256--8265, 2019.

\bibitem{xie2021zsdyolo}
Johnathan Xie and Shuai Zheng.
\newblock Zsd-yolo: Zero-shot yolo detection using vision-language
  knowledgedistillation, 2021.

\bibitem{xu2021simple}
Mengde Xu, Zheng Zhang, Fangyun Wei, Yutong Lin, Yue Cao, Han Hu, and Xiang
  Bai.
\newblock A simple baseline for zero-shot semantic segmentation with
  pre-trained vision-language model.
\newblock {\em arXiv preprint arXiv:2112.14757}, 2021.

\bibitem{ye2019unsupervised}
Mang Ye, Xu Zhang, Pong~C Yuen, and Shih-Fu Chang.
\newblock Unsupervised embedding learning via invariant and spreading instance
  feature.
\newblock In {\em Proceedings of the IEEE/CVF Conference on Computer Vision and
  Pattern Recognition}, pages 6210--6219, 2019.

\bibitem{zareian2021open}
Alireza Zareian, Kevin~Dela Rosa, Derek~Hao Hu, and Shih-Fu Chang.
\newblock Open-vocabulary object detection using captions.
\newblock In {\em Proceedings of the IEEE/CVF Conference on Computer Vision and
  Pattern Recognition}, pages 14393--14402, 2021.

\bibitem{zhao2017pyramid}
Hengshuang Zhao, Jianping Shi, Xiaojuan Qi, Xiaogang Wang, and Jiaya Jia.
\newblock Pyramid scene parsing network.
\newblock In {\em Proceedings of the IEEE conference on computer vision and
  pattern recognition}, pages 2881--2890, 2017.

\end{thebibliography}
}
\end{document}


\pagestyle{headings}
\mainmatter
\def\ECCVSubNumber{4879}  

\title{Appendix for "Open-world Semantic Segmentation via Contrasting and Clustering Vision-Language Embedding"} 

\titlerunning{ViL-Seg}
%
\author{Quande Liu\inst{1} 
\email{qdliu@cse.cuhk.edu.hk}\and
Youpeng Wen\inst{2} \and
Jianhua Han\inst{3} \and
Chunjing Xu\inst{3} \and
Hang Xu\inst{3} \and
Xiaodan Liang\inst{2}\thanks{Corresponding author.}}
\authorrunning{Q. Liu et al.}
%
\institute{The Chinese University of Hong Kong\\ \email{qdliu@cse.cuhk.edu.hk}\and
Shenzhen Campus of Sun Yat-sen University \\
\email{wenyoupeng0@outlook.com, xdliang328@gmail.com}\and
Huawei Noah's Ark Lab\\
\email{\{hanjianhua4,xuchunjing,xuhang\}@huawei.com}
}
\maketitle


\section{Model Architecture}
We follow the popular vision-language training pipeline (e.g., CLIP~\cite{radford2021learning}) to adopt transformer architecture  for the image encoder (ViT-B/16) and text encoder in our ViL-Seg framework. Details of the architecture parameters for the two encoders and the clustering head in our method are summarized in Table~\ref{tab:model_architecture}.

\begin{table}[h]
\centering
\vspace{-2mm}
\caption{Details of the architecture parameters of ViL-Seg model.}
\label{tab:model_architecture}
\scalebox{0.9}{
\begin{tabular}{clc} 
\hline
\hline
\multicolumn{2}{c}{Architecture parameter} &\multirow{1}{*}{Value}\\
\hline
\multirow{5}{*}{Image encoder} &\#Input resolution & 224$\times$224\\
&\#Layer & 12 \\
&\#Width & 768 \\
&\#Head & 12 \\
&\#Embedding dimension &512\\
\hline
\multirow{4}{*}{Text encoder} &\#Layer & 12\\
&\#Width & 512\\
&\#Head & 8 \\
&\#Embedding dimension &512\\
&\# Prompt & ``a photo of a []"\\
\hline
\multirow{2}{*}{Clustering head} &\#Cluster number &20\\
&\#Receptive field &1$\times$1\\
\hline
\hline
\end{tabular}
}
\vspace{-1mm}
\end{table}


\section{Details of Training Hyperparameters}
For cross-modal contrastive loss, we follow CLIP~\cite{radford2021learning} to set the temperature in the softmax function as a learnable parameter initialized as 0.07. In vision-based contrastive loss, we crop 6 local patches for each of input image, and the embedding dimension of the projection layer before computing $L_{vision}$ is 2048. The whole framework is trained end-to-end using Adam optimizer, with the momentum hyperparameter $\beta_1$ and $\beta_2$ set as 0.9 and 0.999 respectively. For the learning rate and weight decay scheduler, we first assign them a base value and then linearly warm them up to the peak value within 4000 iterations. The base values of learning rate and weight decay are both 0, and their peak values are 5e-4 and 0.04 respectively. We totally trained 30 epochs with batchsize of 48 $\times$ 16 = 768 using 48 Tesla V100 16GB. Table~\ref{tab:training_hyperparameter} summarizes the common hyperparameters for ViL-Seg training.

\begin{table}[h]
\centering
\vspace{-2mm}
\caption{Details of the hyperparameters for ViL-Seg training.}
\label{tab:training_hyperparameter}
\scalebox{0.9}{
\begin{tabular}{lcc} 
\hline
\hline
\multicolumn{1}{c}{Hyperparameter} &\multirow{1}{*}{Value}\\
\hline
\multirow{1}{*}{Optimizer} &Adam\\
\multirow{1}{*}{Adam $\beta_1$} &0.9\\
\multirow{1}{*}{Adam $\beta_2$} &0.999\\
Initial temperature in $L_{cross}$ &0.07\\ 
Number of local patches in $L_{vision}$ &6\\
Projection layer dimension in $L_{vision}$ &2048\\
\multirow{1}{*}{Peak learning rate} &5e-4\\
\multirow{1}{*}{Peak weight decay} &0.04\\
\multirow{1}{*}{Warm-up iterations} &4000\\
\multirow{1}{*}{Batch size} &48$\times$16=768\\
\multirow{1}{*}{Gpu number} &48\\
\multirow{1}{*}{Epochs} &30\\
\hline
\hline
\end{tabular}
}
\vspace{-1mm}
\end{table}
\section{More Visualization for Segmenting Open-world Object Categories}
\begin{figure*}[tb]
		\begin{center}
\includegraphics[ width=0.95\linewidth]{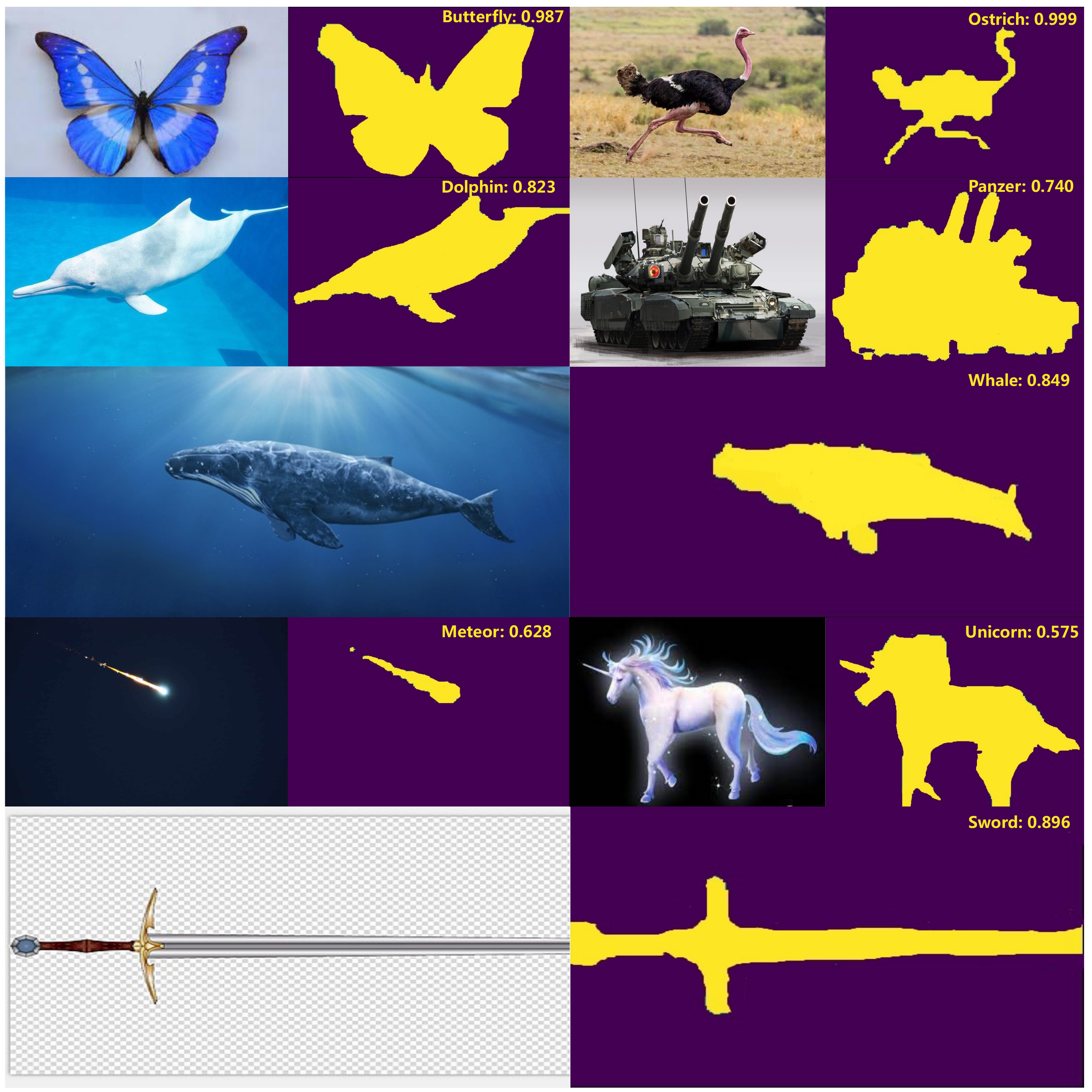}
		\end{center}
		\vspace{-4mm}
		\caption{By purely utilizing the image-caption pairs from the Internet (without using any data with dense annotations), ViL-Seg is able to segment various object categories in the open world even though they are never labeled in existing segmentation datasets.}
		\vspace{-5mm}
		\label{fig:intro_figure}
\end{figure*}

Fig.~\ref{fig:intro_figure} visualizes the segmentation results of ViL-Seg on some rare object categories existing in open-world. Owing to the tremendous data resources on the Internet which contain diverse semantic information of various object categories, the ViL-Seg can accurately segment different objects and predict the correct class name, even though they are never labeled in existing segmentation datasets (e.g., Meteor or sword). These results highlight the advantages of our attempt to purely utilize the vision-language pairs naturally existing on the Internet to learn segmentation models, which can not only improve the model generalizability to novel object categories, but further reduce the human efforts on data labeling to the greatest extent.

\section{Code Release}
The code of ViL-Seg belongs to company asset, and we have submitted an formal application to the company to open source the code upon the release of this paper.

\clearpage
{\small
\bibliographystyle{ieee_fullname}
\bibliography{egbib}
}